\documentclass[letterpaper, 10 pt, journal]{ieeetran}

\IEEEoverridecommandlockouts                              

\usepackage{color}
\usepackage{soul}

\title{
   Progressive-Resolution Policy Distillation: Leveraging Coarse-Resolution Simulations for Time-Efficient Fine-Resolution Policy Learning
}

\author{Yuki Kadokawa$^{1}$, Hirotaka Tahara$^{1,2}$, Takamitsu Matsubara$^{1}$,~\IEEEmembership{Member,~IEEE,}
\thanks{
    This work was supported by JST Moonshot Research and Development, Grant Number JPMJMS2032. 
    \textit{(Corresponding author: Yuki Kadokawa.)} 
    $^{1}$ Nara Institute of Science and Technology, Nara 630-0192, Japan.
    $^{2}$ Kobe City College of Technology, Hyogo 651-2194, Japan.
    kadokawa.yuki@naist.ac.jp, h-tahara@kobe-kosen.ac.jp, takam-m@is.naist.jp
}%
}

\usepackage{setspace}
\usepackage{comment}
\usepackage{subfigure}
\usepackage{booktabs}
\usepackage{graphicx}
\usepackage{amsmath}
\usepackage{amsfonts}
\usepackage{algorithmic}
\usepackage{siunitx}
\usepackage{bm}
\usepackage[space, compress]{cite}
\usepackage[ruled,vlined]{algorithm2e}
\usepackage{algorithm2e,setspace}
\usepackage{textcomp}
\usepackage{mathcomp}
\usepackage{tabularx}
\usepackage{here}
\usepackage[colorlinks=true, linkcolor=blue, citecolor=blue, urlcolor=blue]{hyperref}

\newcommand{\tabref}[1]{\hyperref[#1]{Table~\ref*{#1}}}
\newcommand{\equref}[1]{\hyperref[#1]{Eq.~(\ref*{#1})}}
\newcommand{\figref}[1]{\hyperref[#1]{Fig.~\ref*{#1}}}
\newcommand{\chapref}[1]{\hyperref[#1]{Section~\ref*{#1}}}
\newcommand{\algref}[1]{\hyperref[#1]{Algorithm~\ref*{#1}}}
\newcommand{\apperef}[1]{\hyperref[#1]{Appendix~\ref*{#1}}}

\begin{document}

\maketitle

\begin{abstract}
In earthwork and construction, excavators often encounter large rocks mixed with various soil conditions, requiring skilled operators. This paper presents a framework for achieving autonomous excavation using reinforcement learning (RL) through a rock excavation simulator. In the simulation, resolution can be defined by the particle size/number in the whole soil space. Fine-resolution simulations closely mimic real-world behavior but demand significant calculation time and challenging sample collection, while coarse-resolution simulations enable faster sample collection but deviate from real-world behavior. To combine the advantages of both resolutions, we explore using policies developed in coarse-resolution simulations for pre-training in fine-resolution simulations. To this end, we propose a novel policy learning framework called Progressive-Resolution Policy Distillation (PRPD), which progressively transfers policies through some middle-resolution simulations with conservative policy transfer to avoid domain gaps that could lead to policy transfer failure. Validation in a rock excavation simulator and nine real-world rock environments demonstrated that PRPD reduced sampling time to less than 1/7 while maintaining task success rates comparable to those achieved through policy learning in a fine-resolution simulation. Additional videos and supplementary results are available on our project page: \url{https://yuki-kadokawa.github.io/prpd/}
\end{abstract}

\def\abstractname{Note to Practitioners}
\begin{abstract}
This paper is motivated by the issue of computation time in excavation simulation using soil particles.
The behavior of real soil is highly complex, and approximating it at high resolution requires enormous computational costs.
Therefore, existing soil simulators have focused on improving simulation accuracy while maintaining reduced computation time.
This paper takes a different approach by focusing on the learning of control policies in excavation simulators and proposes a framework for reducing calculation time in such use cases.
In this framework, a control policy is first learned in a low-resolution simulation, significantly reducing computation time.
The learned policy is then transferred to a high-resolution simulation for retraining, thereby achieving an overall reduction in simulation time.
Furthermore, to enable robust policy transfer across different resolutions, this paper discusses a stable policy distillation scheme and insights into resolution design.
This approach enables the development of autonomous excavation systems without relying on expensive real-world data collection, improving the scalability and adaptability of autonomous excavation.
Simulation experiments suggest that this framework significantly reduces training time compared to conventional policy learning approaches.
However, real-world validation has so far been limited to simple excavation robots.
Future research will explore applications to excavators and other machinery more suitable for real-world operations.
Although this paper focuses on autonomous excavation, the proposed approach can also be extended to environments where increased simulation resolution critically impacts computation time, such as liquid and soft object manipulation.
\end{abstract}



\begin{figure}[t]
    \centering
    \includegraphics[width=0.99\columnwidth]{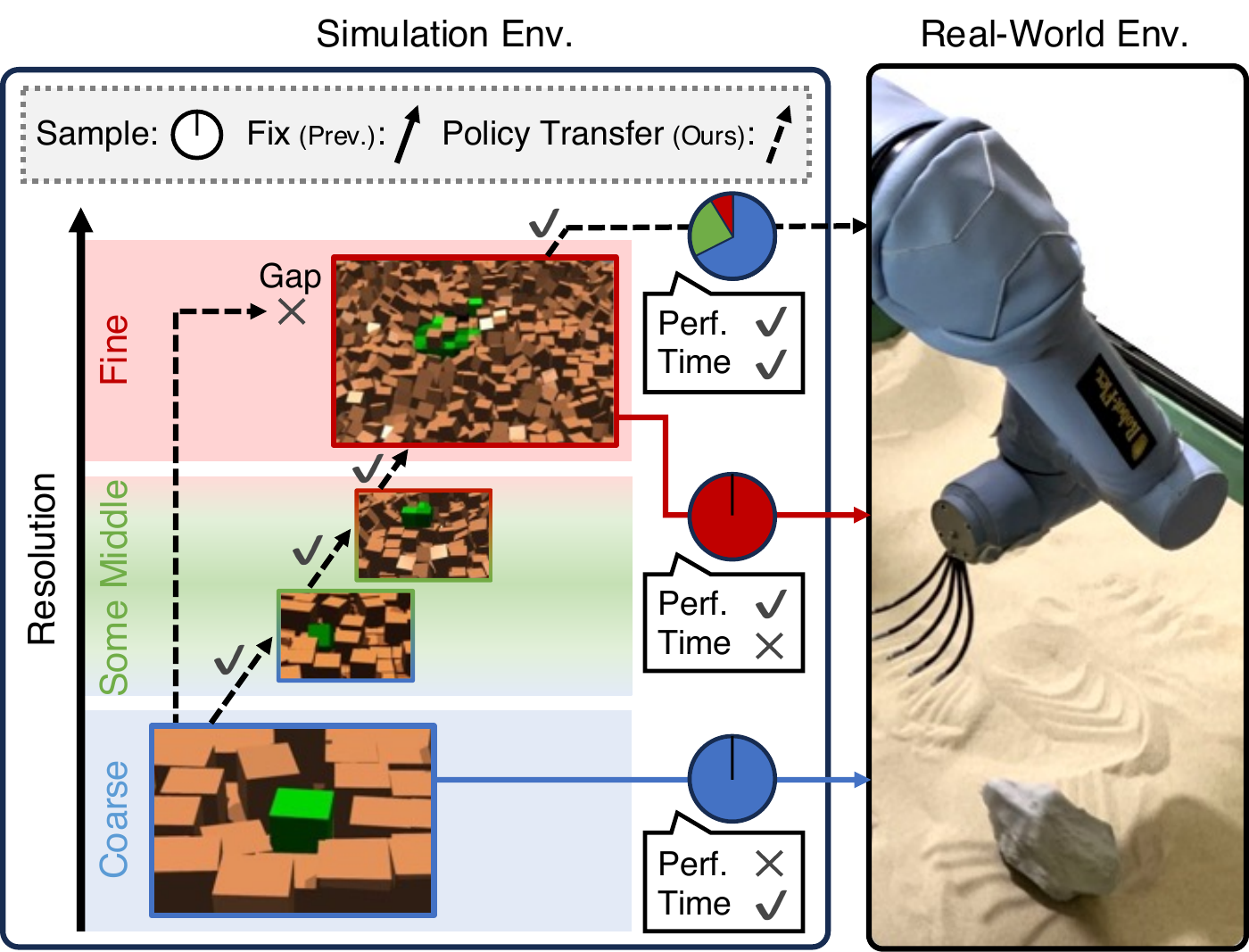}
    \caption{
        Overview of proposed framework:
        Fine-resolution simulations yield high policy performance but require long learning times, while coarse-resolution simulations allow for quick learning but perform poorly in sim-to-real transfer. Our framework starts with coarse-resolution simulations for quick learning and progressively transfers policies to fine-resolution simulations. Progressive resolution shift with conservative policy transfer is applied to avoid large domain gaps that could lead to policy transfer failure. This approach balances learning time with real-world performance.
    }
    \label{fig:proposed_overview}
\end{figure}

\section{Introduction}

    \IEEEPARstart{A}{utonomous} excavation has been studied to meet the rising demand for earthwork and construction \cite{soil_excavation_realOnly_GPmodelMPCPolicyLearning,soil_formula_1,auto_excavation_mpc}. In excavation, especially in mountainous areas, quarry sites, and construction sites, large rocks are often mixed with sand, gravel, and pebbles in the soil \cite{soil_rock_other_soil,rock_excavation_complex_soil}. Operating excavators must have advanced operational expertise to find, collect, and transport these rocks \cite{truck_rock,excavation_is_hard}. For rock excavation, excavators must efficiently lift and move rocks using a bucket. This involves non-grasping bucket motion while considering forces from rock-soil-bucket interactions, a task only skilled operators can perform \cite{rock_excavation_humanBetter, survey_auto_excavation2}. Due to the varied shapes and sizes of rocks, automating rock excavation with simple autonomous control policies is challenging \cite{rock_excavation_realOnly_GPmodelNoPolicyLearning, survey_auto_excavation}.

    Reinforcement learning (RL) is potentially useful to learn optimal actions from diverse state-action transitions caused by various rocks \cite{drl_servey,drl-door,drl_assembly}. RL learns control policies from interaction samples between a robot and the environment. Thus, simulators are commonly used since real-world sampling requires significant manual efforts to set up various types of soil and objects \cite{excavator-sim2real,excavator-sim2real-geometric}. In previous simulation work, computational mechanics and particle methods have been used. Computational mechanics, which discretize terrain into elements, are suitable for simple terrains but struggle with complex terrains involving rocks or nonlinear phenomena \cite{excavator-sim2real,excavator-sim2real-geometric}. Conversely, particle methods model materials as individual particles and simulate their interactions, effectively reproducing complex phenomena like friction, collisions, and soil deformation \cite{vortex-DRL-sim2sim,excavator_sim_particleLikeSim}. This makes particle methods suitable for simulating complex terrains that include rocks.

    While particle-based simulations can be accurate and realistic, they face challenges in computation time for our RL application, particularly as {\it{spatial resolution}} improves. This resolution can be defined by the particle size/number in the whole soil space, with smaller particle sizes (more amounts of particles) allowing for a more detailed particle representation of the soil space and obtaining a finer resolution. Fine resolution leads to complex particle interactions, making it impractical to collect the vast samples needed for RL in a reasonable calculation time \cite{excavator_complex_vortex, PowderWeighingSim2Real}. Ideally, the resolution should be coarsened until the simulator is sufficiently fast; however, simulation deviates from real-world behavior, utilizing the learned policies ineffective in real-world environments.
    Our focus is on addressing RL problems where there is a trade-off between resolution-dependent calculation time and sample quality, aiming to reduce the computation time required for policy learning.

    As shown in \figref{fig:proposed_overview}, we explore using policies developed in coarse-resolution simulations for pre-training in fine-resolution simulations, even though these policies cannot work in real-world settings. This approach potentially reduces sampling time compared to using only fine-resolution simulations since coarse-resolution simulations can be calculated in a shorter time. 
    However, behaviors in different resolution simulations typically deviate due to domain gaps, making it challenging to utilize coarse-resolution simulations for pre-training. 
    Therefore, by transferring policies progressively through some middle resolutions and by transferring policies gradually while checking the stability of the policy updates instead of transferring policies all at once, we could effectively transition these policies to simulations that closely resemble real-world settings.

    We propose a novel policy learning framework called Progressive-Resolution Policy Distillation (PRPD) for achieving time-efficient policy learning of fine-resolution simulations by utilizing coarse-resolution simulations effectively.
    PRPD progressively improves simulation resolution while learning and transferring policies at each stage.
    PRPD uses a conservative policy transfer scheme to regularize policies, stabilizing policy transfer across different resolution simulations having different behaviors. 
    We demonstrated the effectiveness of PRPD by constructing a variable-resolution rock excavation simulator using Isaac Gym \cite{isaacgym}. 
    This paper demonstrated a 7-fold improvement in learning time efficiency (PRPD: 90 minutes, policy learning in the finest-resolution simulation: 600 minutes), resulting in an 8-hour time gap. This difference is likely to become more significant with increasing state-action space complexity or task difficulty.
    Furthermore, in evaluations of the learned policies within a real-world environment containing nine types of rocks, PRPD achieved rock excavation in approximately \SI{90}{\%} of scenarios.

    As an impact of this study, achieving realistic simulations of complex environments, such as excavation tasks, remains challenging. While advances in computational resources and simulation technologies may alleviate simulation time issues, simulating real-world environments accurately remains prohibitively costly and will still face resource limits. This study offers insights into addressing these challenges by improving policy-learning efficiency.

    The following are this paper's main contributions:
    \begin{itemize}
        \item It proposes a novel policy learning framework, PRPD, which enables time-efficient policy learning by progressively increasing simulation resolution and scheduling tasks from coarse to fine based on policy performance.
        \item While the underlying policy optimization and distillation mechanisms are based on existing methods, the novelty of the proposed framework lies in integrating them into a unified structure that enables stable and progressive policy transfer across simulation resolutions. This framework effectively achieves the core novelty of progressive policy distillation across varying simulation resolutions for time-efficient policy learning.
        \item The effectiveness of PRPD is demonstrated in a complex particle-based excavation task, achieving a 7-fold reduction in total learning time compared to fixed-resolution training, while maintaining comparable task performance and enabling sim-to-real policy transfer in diverse real-world environments.
    \end{itemize}

\section{Related Works}
    \label{s:related_works}
    
    \subsection{Learning Excavation Policy}

        This section describes several methods for acquiring control policies for a robotic excavation task with soil and rocks.
        Previous works have proposed the following three approaches.
        
        \textbf{Learning in Real-world:}
            Learning rock-excavation tasks in real-world environments requires extensive sampling time. These works have modeled soil and rock behavior from limited samples to learn control policies \cite{rock_excavation_realOnly_GPmodelNoPolicyLearning,soil_excavation_realOnly_GPmodelMPCPolicyLearning}. However, accurately modeling rock behavior requires interaction samples for various rock shapes. The vast number of combinations makes it difficult to apply this method to multiple shapes due to sampling costs, so previous research has only obtained policies for a single rock shape.
            
        \textbf{Learning in Simulation by Computational Mechanics:}
            These works simulate soil behavior in response to bucket movements using computational mechanics and collect learning samples from model interactions \cite{excavator-sim2real,excavator-sim2real-geometric}. The finite element method divides soil into small elements and solves for displacement, stress, and strain. Deformation analysis uses soil mechanics equations to analyze deformation and failure. 
            These works are suitable for simulating simple terrain because they numerically discretize and simulate terrain, but they have difficulty accurately reproducing complex terrain involving rocks and nonlinear phenomena.

        \textbf{Learning in Simulation by Particle Method:}
            These works adopt the particle method that models materials as individual particles and directly simulates the interactions between them \cite{particle_method_sim_meshFree,ImitationExcavation}. 
            This enables detailed simulation of rock and soil behavior, and approximates complex physical phenomena such as inter-particle friction, collision, and soil deformation separately.
            However, calculating interactions between multiple particles is time-consuming, making it difficult to collect many learning samples. While imitation learning applications exist with few samples, these methods have not been trained in comprehensive environments and can only be applied in limited situations.

        This paper aims to establish learning policies applicable to various rock shapes by utilizing a particle-based simulator capable of achieving this aim. To address the sampling time challenge, we propose a novel policy learning framework that enables the short-time collection of learning samples from various rock shapes, thereby achieving rock excavation for multiple shapes.

    \subsection{Excavation Simulator by Particle Method}

        Excavation simulators with a large number of particle interactions require extensive calculation time \cite{particle_method_sim_meshFree,excavator_sim_particleLikeSim}. 
        The following paragraphs describe previous works that employed three approaches to accelerate calculations and comparisons with our developed simulator.

        \textbf{Static Adjustment of Resolution:}
            To reduce the calculation time for particle interactions, which make up the majority of simulation calculations, efforts have focused on generating environments limited to specific work areas and approximating many soil particles with fewer macro particles \cite{PowderWeighingSim2Real,isaacgym_excavation}. 
            These strategies have decreased the necessary memory resources and sampling time. However, there are limits to how much the resolution can be reduced without diverging from real-world behavior.
        
        \textbf{Dynamic Adjustment of Resolution:}
            Some works accelerate simulations by dynamically changing the resolution \cite{vortex-DRL-sim2sim,excavator_complex_vortex}. They dynamically estimate the work area, splitting particles in fine-resolution regions and merging particles in coarse-resolution areas, thereby reducing calculation time. While this approach achieves acceleration, focusing fine resolution only in the work area still requires significant resources and calculation time, making it unsuitable for the extensive sample collection needed in RL.
        
        \textbf{Parallel Calculation of Particles:}
            Recently, frameworks like Isaac Gym and Isaac Sim have been developed to rapidly simulate robot environments using GPUs, significantly accelerating sampling time \cite{isaacgym,isaacsim}. In excavation environments, these achieve faster calculations by parallelizing the calculation of multiple particle interactions. However, generating a vast number of particles within the simulation requires substantial GPU memory, making it difficult to create parallel environments that can further reduce sampling time \cite{isaacgym_excavation}.

        \textbf{Our Developed Simulator:}
            The simulator developed in this paper incorporates equivalents of the three acceleration techniques of the previous works: (1) approximating soil with micro particles, (2) generating only the work area for fine-resolution simulation, and (3) using Isaac Gym for GPU acceleration. The details of our simulator are described in \chapref{sec:sim_detail}.

        Despite these advancements in simulation acceleration, they remain insufficient for the vast sample collection required in RL \cite{vortex-DRL-sim2sim,excavator_complex_vortex}. RL needs millions of samples, making traditional simulators challenging for policy learning \cite{drl_servey,drl-door}. The goal of this paper is not to develop realistic simulations but to learn policies of realistic simulations. To shorten sampling time, we use both fine-resolution simulations that take longer to calculate and extremely coarse-resolution simulations that require less calculation time.

    \subsection{Sample Efficient Reinforcement Learning Methods}
    \label{related_work:sample_efficient_RL}
        
        Several methods have been proposed to improve sample efficiency in reinforcement learning, each targeting different problem settings with distinct assumptions and trade-offs.
        
        \textbf{Model-based RL:}
            This approach utilizes a learned model of the environment's dynamics to learn policies without direct environment interaction \cite{model_base_rl, model_based_RL, model_based_RL2}. While highly sample-efficient in principle, it is susceptible to performance degradation caused by inaccuracies in the learned model, which makes it less reliable in high-dimensional or partially observable domains. Consequently, applying this approach to environments with complex or hard-to-model dynamics remains challenging.
        
        \textbf{Representation Learning:}
            Representation learning aims to extract informative features from raw observations to improve policy learning efficiency \cite{RL_autoencoder, contrastive_rl, RL_ensemble_auxiliary}. It enhances sample efficiency by enabling more effective use of each observation, particularly in high-dimensional or partially observable environments. However, it often requires additional training objectives and network modules, which increase computational complexity and may demand task-specific tuning.
            In a related direction, data augmentation improves sample diversity by transforming observations and has shown notable success in visual RL settings \cite{augmentation_rl, data_augmentation_RL, data_augmentation}. Nevertheless, its effectiveness depends heavily on the choice of transformation and can be limited in tasks where observations are closely coupled with the underlying dynamics.

        \textbf{Imitation Learning:}
             This approach improves sample efficiency by leveraging expert demonstrations to guide policy learning, thereby reducing the need for extensive exploration and environment interaction \cite{DeepImitation_LSTM_excavation_soil, IL, imitation_learning}. In particular, learning from small demonstration datasets or inferring reward functions from limited supervision can significantly accelerate early-stage training. However, the effectiveness of this approach heavily depends on the quality and diversity of the demonstration data, and it often struggles to generalize beyond the behaviors observed in expert trajectories.
        
        \textbf{Sample-Efficient RL without External Supervision:} 
            Off-policy RL methods, such as Soft Actor-Critic (SAC), improve sample reuse by storing and replaying past interactions from a buffer \cite{SAC, offpolicy_RL}. 
            Also, ensemble learning improves sample efficiency by leveraging multiple models to enhance model generalization \cite{RL_ensemble_auxiliary, ensemble_RL, ensemble_learning}. 
            Although sample-efficient, they typically require large numbers of gradient updates and substantial memory for storing transitions, which may reduce time efficiency.
            Unlike methods that rely on external supervision, these approaches operate solely on state-action interaction data and require no additional annotation, reward shaping, or auxiliary training objectives.

        Various methods have improved RL's ``sample efficiency'' by extracting more information from each sample or increasing update times to maximize sample utilization. However, these approaches often come with computational overhead, such as increased calculation time and resource demands by adding learning components, which can compromise ``time efficiency,'' the overall training time for policy learning.

    \subsection{Comparison with Curriculum Reinforcement Learning} 
        Curriculum Reinforcement Learning (CRL) is an RL method that starts with simple tasks and gradually increases task difficulty \cite{CRL}. This approach builds on the initial learning experiences from simple tasks and gradually adapts the policies for more complex tasks. Eventually, this progressive increase in difficulty results in a reduction in the number of samples required compared to learning solely from complex tasks \cite{curriculum_learning_Visuomotor}.

        In previous CRL approaches, the learning process is typically organized using a curriculum based on task difficulty, while the simulator resolution remains fixed. For example, some methods restrict the range of goals or initial states during early training and gradually expand this range as the policy stabilizes, thereby guiding the agent toward a policy with better generalization capabilities \cite{CRL_autoGoal1, CRL_autoGoal2}. Other approaches begin training in simplified environments where external disturbances or noise are intentionally removed, and incrementally introduce real-world complexity to adjust the difficulty of learning \cite{CRL_noise1, CRL_noise2}. In addition, reward shaping has proven effective as a curriculum design strategy, where the reward function is gradually modified. It typically starts with easily obtainable rewards to promote exploration during early training and then transitions to the original reward structure \cite{CRL_reward1, CRL_reward2}. Thus, previous CRL approaches primarily rely on task difficulty as the basis for designing the learning curriculum.
        
        Our approach can be interpreted as a type of CRL, as it also involves gradually changing the tasks from simple to complex. However, unlike previous CRL approaches, our method aims to reduce the overall learning time required for policy training in simulation, rather than minimizing the number of samples used for training, which is the typical objective in common CRL. To the best of our knowledge, this is the first approach to treat differences in the spatial resolution of complex simulations as distinct tasks within a curriculum framework.

\section{Preliminaries}
    \label{preliminaries}

    \subsection{Reinforcement Learning}
    \label{preliminaries:rl}

        Reinforcement Learning (RL) is a framework where an agent learns an optimal policy through interaction with the environment \cite{RL}. Typically, RL is formulated as a Markov Decision Process (MDP) $\mathcal{M}$. An MDP is defined by the state space $\mathcal{S}$, action space $\mathcal{A}$, state transition probability $P(s'|s, a)$, and reward function $R(s, a)$, represented as $\mathcal{M}=(\mathcal{S}, \mathcal{A}, P, R)$. The agent selects an action $a \in \mathcal{A}$ in the current state $s \in \mathcal{S}$ and, as a result, observes the next state $s' \in \mathcal{S}$ and reward $r \in \mathbb{R}$. The goal in an MDP is to find a policy $\pi(a|s)$ that maximizes the expected cumulative reward $J(\pi)$, given by:
        \begin{equation}
            J(\pi) = \mathbb{E}_{\pi} \left[ \sum\nolimits_{t=0}^{\infty} \gamma^t r_t \right],
        \end{equation}
        where $\gamma \in [0, 1]$ is the discount factor.

        To maximize the expected cumulative reward $ J(\pi) $, policy gradient methods are used, where the objective function is defined as:
        \begin{equation}
            J(\pi) = \mathbb{E}_{\pi} \left[ \sum\nolimits_{t=0}^{\infty} A(s_t, a_t) \right],
        \end{equation}
        where $ A(s, a) = Q(s, a) - V(s) $ is the advantage function. The value function $ Q(s, a;\psi) = \mathbb{E}_{\pi} \left[ \sum_{t=0}^{\infty} \gamma^t r_t \big| s_0 = s, a_0 = a \right] $ represents the expected cumulative reward when action $ a $ is taken in state $ s $. The value function $ V(s;\phi) = \mathbb{E}_{\pi} \left[ \sum_{t=0}^{\infty} \gamma^t r_t \big| s_0 = s \right] $ represents the expected cumulative reward starting from state $ s $ and following policy $ \pi(s, a;\theta) $.
        These functions are typically approximated using function approximators such as neural networks. Here, $ \theta $, $ \phi $, and $ \psi $ denote the parameters of the policy $ \pi $, the value functions $ V, Q$, respectively.

        A representative method based on policy gradient methods is proximal policy optimization (PPO) \cite{PPO}. PPO learns an optimal policy by iteratively collecting samples $(s, a, r, s')$ with the policy $\pi$ into a rollout buffer $\mathcal{D}$, and updating the policy using these samples. The number of such iterations is denoted as $i$, and the number of parameter updates is denoted as $k$. PPO updates the policy by minimizing the following loss function:
        \begin{equation}
            \label{eq:ppo_loss}
            \begin{aligned}
                \mathcal{L}_{\theta, \phi}^{\pi, V}  = \ & \mathbb{E}_{\mathcal{D}_i} \left[ \min \{ \rho A_k(s, a), \text{clip}(\rho, 1 - \epsilon, 1 + \epsilon ) A_k(s, a) \} \right. \\
                & \left. - c_1 (V_k(s) - \hat{R})^2 + c_2 \pi_k(a|s) \log \pi_k(a|s) \right] ,
            \end{aligned}
        \end{equation}
        where $\rho$ represents the importance sampling ratio, and $\epsilon$ is a hyperparameter controlling the clipping range. The parameters $c_1$ and $c_2$ control the weights of the value function loss and the entropy regularization, respectively.
        $\hat{R}$ is the cumulative reward calculated from the samples in $\mathcal{D}$ \cite{PPO}.

    \subsection{Conservative Policy Update in Reinforcement Learning}
        In RL, there are various errors such as function approximation errors and observation noise, and attempts have been made to learn policies that are robust to these errors.
        Conservative Policy Iteration (CPI) is one example, which is designed to achieve more stable policy updates by relaxing the policy updates scheme \cite{DCPI,CAC}.

        In CPI, the policy update is modified to make it more conservative by introducing a mixing coefficient $\alpha$. The new policy $\pi_{k+1}$ is a mixture of the current policy $\pi_k$ and the greedy policy $\mathcal{G}(Q_k)(s) = \arg\max_{a} Q_k(s, a)$:
        \begin{equation}
            \label{eq:mi_combine_policy}
            \pi_{k+1}(s,a) \leftarrow (1 - \alpha_{k+1}) \pi_k(s,a) + \alpha_{k+1} \mathcal{G}(Q_k) (s),
        \end{equation}
        where $0 \leq \alpha_{k+1} \leq 1$. This approach stabilizes policy learning by mitigating abrupt policy changes.
        
        CPI comes with strong theoretical guarantees that ensure the policy value improves monotonically under certain conditions; if the function approximation error is bounded and the mixing coefficient $ \alpha_k $ is chosen appropriately, the expected value of the policy is guaranteed to improve. Specifically, the mixing rate can be selected as:
        \begin{equation}
            \label{eq:CPI-update}
            \alpha_{k+1}= \frac{(1 - \gamma)}{4 R} \sum_{s \in \mathcal{S}} d^{\pi_{k}}_{\mu}(s) \sum_{a \in \mathcal{A}}\pi_{k}(a|s)A_k(s,a),
        \end{equation}
        where $d^{\pi_{k-1}}_{\mu}(s)$ is the discounted state visitation distribution under policy $ \pi_{k-1} $ starting from initial state distribution $ \mu $, $ R $ is the maximum reward \cite{DCPI}.

\section{Progressive-Resolution Policy Distillation}
    
    We propose a novel policy learning framework called Progressive-Resolution Policy Distillation (PRPD) to achieve time-efficient policy learning of fine-resolution simulations by utilizing coarse-resolution simulations effectively. 
    An overview of the proposed PRPD framework is shown in \figref{fig:proposed_overview}. 
    PRPD progressively improves simulation resolution and transfers policies at each stage.
    By transferring policies progressively through some middle resolutions and by transferring policies gradually while checking the stability of the policy updates instead of transferring policies all at once, we effectively transition these policies to simulations that closely resemble real-world settings. PRPD incorporates iterations of the following two steps:
    1) Resolution Scheduling and 2) Policy Learning.
    The following sections outline the details of the learning steps. The pseudo-code is provided in Algorithm~$1$.

    \textbf{Simulator Assumption:}
        To enable policy transfer, our framework follows some simulator assumptions. Policies are designed consistently across different resolutions, even though different simulators typically vary in observations and actions. We assume that resolution-affected elements do not impact observations or actions. The policies' objectives remain the same, with only the resolution differing, so the reward function is kept identical. This framework supports policy learning in high-resolution simulations, assuming that behavior is less accurate at coarser resolutions and becomes finer as resolution improves.
        Each resolution level corresponds to a distinct simulation environment configured with different resolution parameters, such as soil particle size. Learning samples are collected independently through interactions between policies and their corresponding environments. We assume that resolution-specific environments can be prepared in advance.

    \textbf{RL Scheme Assumption:}
        PRPD assumes an actor-critic structure for the RL scheme, it requires policy $\pi$ and value function $Q$ for estimating $\alpha$. Applicable learning methods include the latest DRLs PPO \cite{PPO} and SAC \cite{SAC}.

\subsection{Resolution Scheduling}
    \label{sec:resolution_schedular}

    \subsubsection{Execution Flow}
        The simulation resolution $\Delta \in \mathbb{R}$ is scheduled based on the progress of policy learning by the resolution scheduler. Specifically, the resolution remains constant until the policy achieves the task, at which point it is progressively improved by $\Delta_{\mathcal{R}}$. The success rate is evaluated across all results of each iteration, with the task achievement defined by meeting the threshold $\tau$. The simulation generator then creates the simulator $\mathcal{M}^{(n)}$ according to the scheduled resolution $\Delta$.

    \subsubsection{Scheduling Setups}
        PRPD creates a simulation environment represented as MDPs $\mathcal{M}^{(1)}, \mathcal{M}^{(2)}, \ldots, \mathcal{M}^{(N)}$.
        We design the resolution interval $\Delta_{\mathcal{R}}$ between $\mathcal{M}^{(n-1)}$ and $\mathcal{M}^{(n)}$ to be sufficiently small so that the policy $\pi^{(n-1)}$ at the previous resolution can be stably transferred to the next policy $\pi^{(n)}$. The resolution scheduler orders multiple simulation environments, allowing the agent to learn tasks progressively.

\begin{algorithm}[t]
    \label{algorithm}
    \footnotesize
    \SetKwData{Left}{left}\SetKwData{This}{this}\SetKwData{Up}{up}
    \SetKwFunction{Union}{Union}\SetKwFunction{FindCompress}{FindCompress}
    \SetKwInOut{Input}{input}\SetKwInOut{Output}{output}
    \begin{minipage}{1.1\textwidth} 
        \caption{\footnotesize Progressive-Resolution Policy Distillation with PPO}
    \end{minipage}
    \SetKwFunction{UpPN}{PolicyUpdate}
    \SetKwFunction{DC}{DataCollect}
    \SetKwProg{Fn}{Function}{:}{\KwRet}
    \hspace{-4mm}\# Set parameters described in \tabref{table:parameter_setting} \\
    \# Schedule resolution progressively $(\Delta_1, \Delta_2, ..., \Delta_N)$ \\
    \For{$ n = 1, 2, ..., N$}
    {
        \# Initialize iteration num $i \! = \! 0$, parameter update num $k \! = \! 0$ \\
        \# Set values $V^{(n)}_k$, $Q^{(n)}_k$, policy $\pi^{(n)}_k$, rollout buffer $\mathcal{D}^{(n)}_i$ \\
        \# Copy network parameters ($\Delta_{n-1}$ to $\Delta_n$ ) \\
        \# Set resolution $\Delta \leftarrow \Delta_n$ and generate simulator $\mathcal{M}^{(n)}$ \\
        \While{until $\tau > \hat{\tau}$}
        {
            \For{$ e = 1, 2, ..., E$}
            {
                \For{$ t = 1, 2, ..., T$}
                {
                    \# Take action $a_t$ from $\pi^{(n)}_k$ in $\mathcal{M}^{(n)}$ \\
                    \# Get observation $s_{t}$, reward $r_t$ \\
                    \# Push $(s_t, a_t,r_t)$ to $\mathcal{D}^{(n)}_i$ \\
                }
            }
            \For{$ k' = k, k+1, ..., k+K$}
            {
                \# Calculate loss: \equref{eq:proposed-all-loss}, update $V^{(n)}_k, Q^{(n)}_k, \pi^{(n)}_k$ \\
            }
            \# update numbers $i=i+1$, $k=k+K$ \\
            \# Check success rate $\tau$ \\
        }
    }
\end{algorithm}

\begin{table}[t]
    \vspace{1mm}
    \caption{
            Learning parameters of PRPD in experiments.
            \label{table:parameter_setting}
    }
    \vspace{-2mm}
    \begin{center}
        \footnotesize
        \begin{tabular}{@{}lp{5.5cm}llll@{}}
            \toprule
            \textbf{Para.} & \textbf{Meaning} & \textbf{Value}  \\ 
            \midrule
            $\alpha_0$ & Scaling coefficient of distillation & 2 \\
            $\gamma$ & Discount factor of RL & 0.99 \\
            $\Delta_{1}$ & Scale of initial resolution [mm] & 70 \\
            $\Delta_{N}$ & Scale of final resolution [mm]& 10 \\
            $\Delta_{\mathcal{R}}$ & Scale of resolution interval [mm] & 10 \\
            $\hat{\tau}$ & Target success rate & 0.95 \\
            $T$ & Number of steps per episode & 128 \\
            $E$ & Number of episodes per iteration & 128 \\
            $K$ & Number of parameter update per iteration & 64$\times$8 \\
            $c_3$ & Weight of distillation loss & 0.5 \\
            $c_4$ & Weight of Q-value loss & 1 \\
            \bottomrule
        \end{tabular}
    \end{center}
\end{table}

\subsection{Policy Learning}    
    
    \subsubsection{Execution Flow}
        An agent collects samples $s, a, r$ from interactions with the simulation and adds them to the rollout buffer $\mathcal{D}^{(n)}$ corresponding to the current resolution. The agent then updates $Q^{(n)}$ and $\pi^{(n)}$ using the RL update scheme. At the end of each episode, the success rate $\tau$ of the policy is evaluated and passed to the resolution scheduler.
        When the resolution scheduler updates the resolution $\Delta$, the simulation environments are updated by the simulation generator.
        Then, the previous value function $Q^{(n-1)}$ and policy $\pi^{(n-1)}$ are copied to that of the current resolution $Q^{(n)}$, $\pi^{(n)}$.

\begin{figure}[t]
    \centering
    \includegraphics[width=0.99\columnwidth]{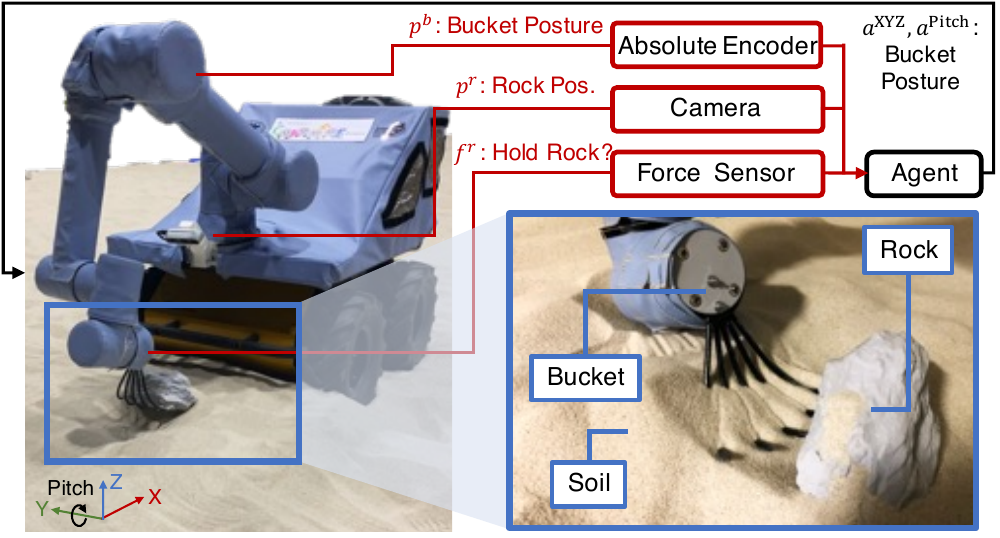}
    \caption{
        Our experimental rock excavation setup:
        The excavator operates a bucket attached to its arm to remove rocks from the soil. Inputs to the control policy include the bucket's posture $p^b$ (position and rotation) from the excavator's absolute encoder, rock coordinates $p^r$ estimated by the camera, and the presence of rocks in the bucket $f^r$ estimated by the force sensor. The output of the control policy is the position $a^{\text{XYZ}}$ and rotation of the bucket $a^{\text{Pitch}}$.
        The fork-shaped bucket is designed to imitate the features of skeleton buckets.
    }
    \label{fig:real_env}
\end{figure}

    \subsubsection{Policy Update with Conservative Policy Transfer}
        
        PRPD approximately utilizes the conservative policy update scheme to stabilize policy transfer between different resolutions.
        This scheme was originally developed for a single environment $\mathcal{M}$ in a previous work \cite{DCPI}. This paper approximately utilizes this approach by assuming that a small resolution interval between environments makes these environments similar to a single environment ($\mathcal{M}^{(n)} \approx \mathcal{M}^{(n-1)}$).        
        Therefore, it can be inferred that the lower the similarity (the larger the change in resolution), the lower the effectiveness, which will be verified in detail in \chapref{ex-MI}.
        For this purpose, we extend the policy linear combination in \equref{eq:mi_combine_policy}. Specifically, whereas \equref{eq:mi_combine_policy} conservatively transfers the greedy policy $\mathcal{G}(Q^{\pi})$, PRPD conservatively transfers the learned previous resolution policy $\pi^{(n-1)}$ as follows:
        \begin{equation}
            \label{eq:proposed_combine_policy}
            \pi_{k+1}^{(n)}(s,a) \gets (1-\alpha_{k+1}) \pi_{k}^{(n)}(s,a) + \alpha_{k+1} \pi^{(n-1)}(s,a).
        \end{equation}
        As shown in \equref{eq:CPI-update}, the maximum reward and stationary distribution $d_{\pi,\mu}$ are theoretically necessary for estimating $\alpha$, but obtaining these parameters in a realistic task is seldom possible. Therefore, we modify \equref{eq:proposed_combine_policy} to be able to infer by mini-batch of deep reinforcement learning (DRL) by following previous works \cite{DCPI,CAC} as:
        \begin{equation}
            \label{eq:proposed-alpha-update}
            \begin{aligned}
                \alpha_{k+1} = \alpha_{0} \ \mathbb{E}_{s \sim \mathcal{D}_{i}^{(n)}} & \big[\mathbb{E}_{a' \sim \pi^{(n-1)}(s)} [Q_{k}^{(n)}(s,a')] \\ 
                & - \mathbb{E}_{a \sim {\pi}_{k}^{(n)}(s)} [Q_{k}^{(n)}(s,{a})] \big], 
            \end{aligned}
        \end{equation}
        where $\alpha_0$ is a scaling coefficient, $\mathcal{D}^{(n)}$ is the rollout buffer of the current resolution, and $k$ is an update number. Estimated $\alpha_{k+1}$ is utilized by clipping to $[0,1]$.
        This weighting scheme increases $\alpha$ when the current policy outperforms the previous policy, thereby making the framework robust to the poor teacher effect \cite{DualPD,DistillingPD} by applying distillation only when it is expected to improve the student policy, as suggested in previous work \cite{cpd}.
        Finally, the loss function of conservative policy update is defined:
        \begin{equation}
            \label{eq:proposed-policy-loss}
            \begin{aligned}
                \mathcal{L}_{\theta^{(n)}}^{\pi^{(n)}} = \ & \mathbb{E}_{(s, a) \sim \mathcal{D}_{i}^{(n)}} \big[ \text{KL} \{ \pi^{(n)}_k (s,a)|| \\
                & (1-\alpha_{k+1}) \pi^{(n)}_k (s,a) + \alpha_{k+1} \pi^{(n-1)} (s,a) \} \big],  
            \end{aligned}
        \end{equation}
        where $\theta^{(n)}$ represents the policy network parameters for the current resolution, and $\text{KL}$ denotes the Kullback–Leibler divergence.  
        To estimate the policy mixture rate $\alpha$, PRPD learns an auxiliary value function $Q^{(n)}$, parameterized by $\psi^{(n)}$, which is not used for PPO updates but exclusively for estimating $\alpha$ across resolutions. The value function $Q^{(n)}(s, a)$ is trained independently from the actor-critic structure via a temporal-difference-based loss defined as:
        \begin{equation}
            \label{eq:proposed-q}
            \begin{aligned}
                \mathcal{L}_{\psi^{(n)}}^{Q^{(n)}} = \ & \mathbb{E}_{(s, a, r, s') \sim \mathcal{D}_{i}^{(n)}} \big[ \{ Q^{(n)}_k (s, a) - \\
                & r + \gamma \mathbb{E}_{a' \sim \pi^{(n)}(s')} [ Q^{(n)}_k (s', a') ]  \}^2 \big] .
            \end{aligned}
        \end{equation}
        Finally, the policy network is updated by combining the PPO loss (\equref{eq:ppo_loss}) with the KL-based loss and the Q-function loss into the following overall loss function:
        \begin{equation}
            \label{eq:proposed-all-loss}
            \mathcal{L}_{\theta^{(n)},\phi^{(n)},\psi^{(n)}} =  \mathcal{L}_{\theta^{(n)}, \phi^{(n)}}^{\pi^{(n)}, V^{(n)}} + c_3 \mathcal{L}_{\theta^{(n)}}^{\pi^{(n)}} + c_4 \mathcal{L}_{\psi^{(n)}}^{Q^{(n)}},
        \end{equation}
        where $c_3$ and $c_4$ are hyperparameters.

\begin{figure}[t]
    \centering
    \includegraphics[width=0.99\columnwidth]{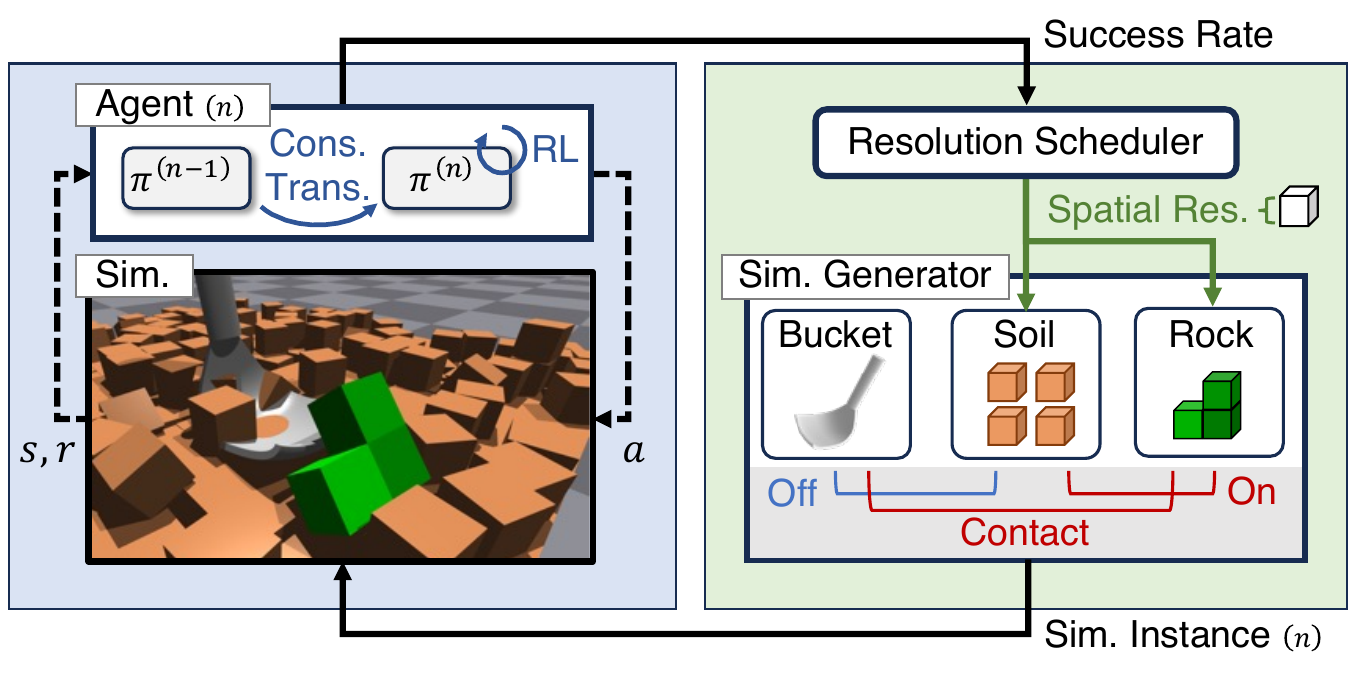}
    \vspace{-1mm}
    \caption{
        Applying PRPD to our rock excavation simulator:
        The resolution scheduler progressively changes the simulation resolution.
        The simulation generator creates the environment (soil, rocks, bucket) at this resolution. At each resolution, agents collect samples and update policies.
    }
    \label{fig:method_to_rockEnv}
\end{figure}

\section{Rock Excavation}
\label{s:problem}
    
    This section outlines the task definition and the simulator developed for learning rock excavation using RL. The objective is to remove various rocks from the ground under different conditions using the bucket. To achieve this, we created both a real-world environment (\figref{fig:real_env}) and a simulation environment (\figref{fig:method_to_rockEnv}). We describe the design of excavation motions and sensors optimized with RL and formalize rock excavation as an RL problem.

\subsection{Task Motions}
    The rock excavation task is defined in three steps: 1) find the rock and move the excavator to it, 2) pick up the rock from the ground, and 3) dump the scooped rock at a designated place. We focus on learning the pickup operation due to the complexity of soil and rock mechanics. To make the learning process more manageable, we observed human operators and designed parameterized motions for inclining and moving the bucket straight. 
    The incline action allows the bucket to insert into the ground at an angle and tilt to hold the rock, while linear bucket movement helps lift the rock and adjust the relative position between the bucket and the rock. Each motion is optimized with RL.

\begin{table}
    \begin{center}
    \caption
    {
        Calculation time per one-step sample in each resolution.
        \label{table:calculation_time_rate_list}
    }
    \footnotesize
        \begin{tabular}{lccccccc}
            \toprule
                Res. $\Delta$ [mm] & 70 & 60 & 50 & 40 & 30 & 20 & 10 \\
                Time [ms] & 0.24 & 0.26 & 0.34 & 0.41 & 0.94 & 2.97 & 5.03 \\
            \bottomrule
        \end{tabular}
    \end{center}
\end{table}

\subsection{Sensors and Observations}
    We designed three key observations for the excavation task: rock position estimated by the camera, bucket force measured by a force sensor, and bucket position and incline degree recorded by the absolute encoder. Rock position is derived from RGB images and the robot's relative position by estimating the image-based center of gravity. 
    The bucket force observation is represented as a binary value: only the vertical force measured by the sensor is used, with a value of 1 if it exceeds a threshold (indicating contact with a rock or the ground) and 0 otherwise (indicating the bucket is in the air). 
    The bucket's incline is limited to the scooping direction to reduce action dimensions and, if the torque limit is exceeded, the arm's movement in that direction is restricted.
    Each observation is recorded once at the end of the agent's action and is used as next agent's observation.

\subsection{Learning Formulation}
    \label{sec:learning_formulation}
    We formulate this rock excavation as an RL problem.
    The observation $s$ is defined from the estimated rock posture (XY position and pitch rotation) $p^{\text{r}}$, the estimated rock force $f^{\text{r}}$, and bucket position $p^{\text{b}}$, $s=[p^{\text{r}}, f^{\text{r}}, p^{\text{b}}]$.
    As supplementary information, soil details (such as appearance and particle count) differ between resolutions and exceed the assumptions of the proposed framework; to enable policy transfer across different resolutions, these details are excluded from observations.
    The action $a$ is defined as $a=[a^{\text{XYZ}}, a^{\text{Pitch}}]$: $a^{\text{XYZ}}$ is the move straight bucket action that represents the relative XYZ position of the bucket and $a^{\text{Pitch}}$ is the inclining bucket action that represents the relative pitch angle of the bucket. Both $a^{\text{XYZ}}$ and $a^{\text{Pitch}}$ are executed simultaneously at every step $t$.
    The reward $r$ is defined from the current rock height $h^{\text{rock}}$, $r=h^{\text{rock}}$.
    The task success is evaluated by determining whether the rocks are above ground height $h^{\text{ground}}$ using the following indicator function: $\mathbb{I}(h^{\text{rock}} > h^{\text{ground}})$.

\begin{figure}[t]
    \centering
    \includegraphics[width=0.99\columnwidth]{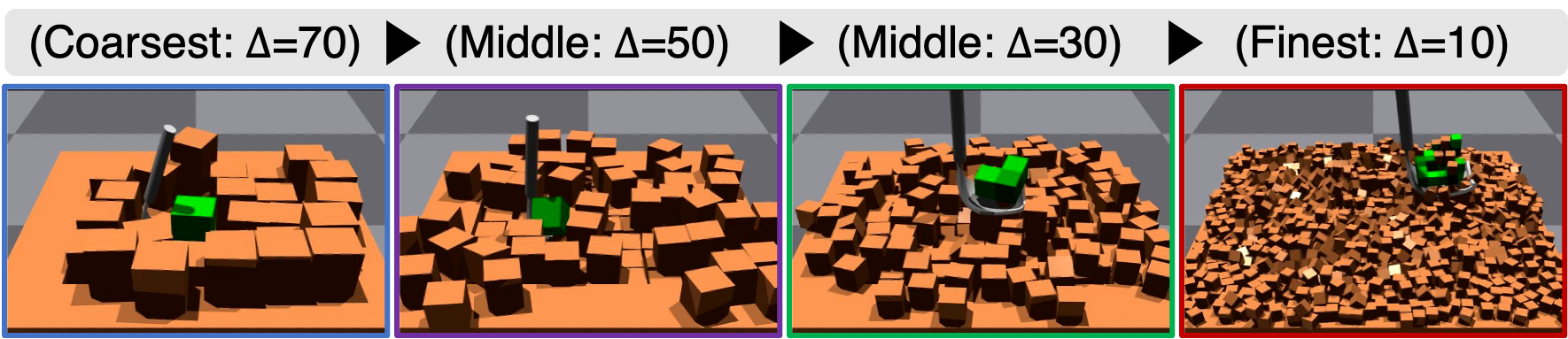}
    \caption{
        Snapshots of simulation environments with each resolution
    }
    \label{fig:sim_resolution_view}
\end{figure}

\subsection{Variable Resolution Simulator}
\label{sec:sim_detail}
    We designed and implemented a variable-resolution simulator for rock excavation tasks using Isaac Gym \cite{isaacgym}. 
    The simulator is presented in \figref{fig:method_to_rockEnv} and enables us to remove the various rocks on various ground types by inclining and moving the bucket. 
    The excavator's whole body model and the process of moving the excavator to the rocks and putting the rocks in other places are excluded. 
    The rock position is obtained directly by the simulation property.
    3D CAD software was utilized to design the bucket.
    The shape of the ground soil particle was assumed to be a box. 
    The bucket was designed to make contact only with the rock, eliminating resistance to particles representing soil; this is for the purpose of expressing that the fork-shaped bucket can move and dig out the soil.

    Our rock-excavation simulator generates environments containing soil and rocks based on the specified simulation resolution $\Delta$. In this paper, we focus on spatial resolution, dynamically adjusting parameters such as soil particle size and count, and rock shape precision. For soil, particles are generated as boxes with side lengths represented by resolution $\Delta$ to match a predefined total soil volume. 
    Rocks are represented as a collection of connected boxes. Specifically, boxes with side lengths represented by resolution $\Delta$ are connected face-to-face until the defined rock volume is reached. To create diverse rock shapes, the faces of the connected boxes are randomly selected.
    
    The calculation times and environment snapshots at different resolutions for our developed simulator are shown in \tabref{table:calculation_time_rate_list} and \figref{fig:sim_resolution_view}, respectively. The calculation time of this simulator improves as resolution $\Delta$ becomes finer due to more soil particles and complex rock shapes, which result in more contact points and higher calculation times, as well as increased GPU memory usage. Consequently, finer-resolution simulations require significant calculation time, making the collection of the vast number of learning samples needed for RL very challenging.

\begin{table}
    \vspace{1.mm}
    \begin{center}
    \caption
    {
        Range of randomized parameters in the rock excavation simulator:
        {\normalfont
        The parameters were used for learning policies with DR and sampled from the uniform distribution.
        ``XY'' and ``XYZ'' denote whether only the horizontal direction or the vertical direction is included, respectively.
        }
        \label{table:domain-parameter-real}
    }
    \footnotesize
        \begin{tabular}{lccc}
            \toprule
                Parameter & w/o DR & min & max \\
            \midrule
                Obs. of Rock pos. noise (XY) $[\textrm{mm}]$ & 0 & -25 & 25 \\
                Obs. of Rock pos. bias (XY) $[\textrm{mm}]$ & 0 & -25 & 25 \\
                Error rate of rock in bucket $[\cdot]$ & 0 & 0.2 & 0.2 \\
                Ground height bias $[\textrm{mm}]$ & 0 & -25 & 25 \\
                Init. bucket pos. bias (XYZ) $[\textrm{mm}]$ & 0 & -300 & 300 \\
                Init. rock pos. bias (XYZ) $[\textrm{mm}]$ & 0 & -30 & 30 \\
                Bucket torque weight (XYZ) $[\cdot]$ & 1 & 0.8 & 1.2 \\
                External force to rock $[\textrm{N}]$ & 0 & 0 & 1 \\
                Friction coefficient $[\cdot]$ & 1 & 0.8 & 1.2 \\
                Total soil mass $[\textrm{kg}]$ & 3 & 2.7 & 3.3 \\
                Total rock mass $[\textrm{kg}]$ & 1 & 0.8 & 1.2 \\
                Total soil volume $[\textrm{mm}^3]$ & $125^3$ & $120^3$ & $130^3$ \\
                Total rock volume $[\textrm{mm}^3]$ & $50^3$ & $45^3$ & $55^3$ \\
            \bottomrule
        \end{tabular}
    \end{center}
\end{table}

\begin{table*}
    \begin{center}
    \caption
    {
        Task Success Rate Comparison at Trained Resolution and at the Finest Resolution ($\Delta=10$)
        \label{table:compare_achieve_corr_delta10}
    }
    \footnotesize
        \begin{tabular}{lccccccc}
            \toprule
                Trained Resolution $\Delta$ [mm] & 70 & 60 & 50 & 40 & 30 & 20 & 10 \\
            \midrule
                Trained Resolution [\%] & $99.4 \pm 2.6$  & $99.1 \pm 3.1$ & $98.9 \pm 2.3$ & $98.3 \pm 1.9$ & $97.5 \pm 5.6$ & $96.9 \pm 7.2$ & $95.8 \pm 6.4$ \\
                $\Delta = 10$ [\%] & $61.1 \pm 8.7$  & $68.0 \pm 8.3$ & $76.0 \pm 9.7$ & $81.3 \pm 5.8$ & $85.0 \pm 7.5$ & $91.1 \pm 8.9$ & $95.8 \pm 6.4$ \\
            \bottomrule
        \end{tabular}
    \end{center}
\end{table*}

\section{Experiments}
    \label{s:exp}
    We conducted experiments to validate the following objectives. PRPD can learn policies in a shorter learning time compared to previous works (\chapref{ex-time}). 
    The conservative policy transfer scheme stabilizes learning in PRPD (\chapref{ex-MI}). 
    Differences in resolution scheduling affect policy learning (\chapref{ex-scheduling}). 
    Sensitivity of PRPD to loss weight scaling (\chapref{ex-loss_scale}). 
    Trade-off between time efficiency and sample efficiency compared to other RL methods (\chapref{ex-time_sample}).
    The policy trained in the simulator can be applied to various real-world environments (\chapref{ex-various_rock}).

\subsection{Common Settings}
    
    The experiments used a Universal Robots UR5e manipulator to control rocks with a bucket. Rock position was estimated using an Intel RealSense Depth Camera D435 and bucket force was measured with a Robotiq FT 300-S Force Torque Sensor. The robot controlled the fingertip pose at approximately 100 Hz. Policy learning performance was evaluated in simulation using an Intel Core i9-9900X CPU and GeForce RTX 3090 GPU. All experiments used RL parameters as shown in \tabref{table:parameter_setting} and the network architecture from \cite{DR-DRL-LSTM}. For real-world evaluation, nine types of rocks with different shapes and sizes were used as shown in \tabref{table:achievement}. 
    Policy learning employed PPO with parallel environments; $E$ environments are simulated in parallel at all resolutions, which is the maximum number of parallel environments achievable on the current PC setup for the highest resolution environment ($\Delta = 10$).
    Advantage $A$ was estimated using the generalized advantage estimation scheme \cite{GAE}, and used in the PPO update described in \equref{eq:ppo_loss}.

    This paper applies a domain randomization (DR) technique \cite{DR-DRL-LSTM, cpd} to learn robust policies for reality gaps between the simulation and the real world. The learned policy is then utilized in real-world environments without additional learning costs.
    We randomized simulation parameters during training as described in \tabref{table:domain-parameter-real}.
    These parameters were uniformly randomized at every episode.

    This experiment evaluates three key aspects. \textbf{Sample number} is defined as the total number of samples collected by the policy. \textbf{Success rate} is defined as the percentage of episodes per iteration that achieve the task, with success defined in \chapref{sec:learning_formulation}. \textbf{Learning time} is defined as the total training time, including both environment interactions and policy optimization.

\begin{figure}[t]
    \centering
    \includegraphics[width=0.99\columnwidth]{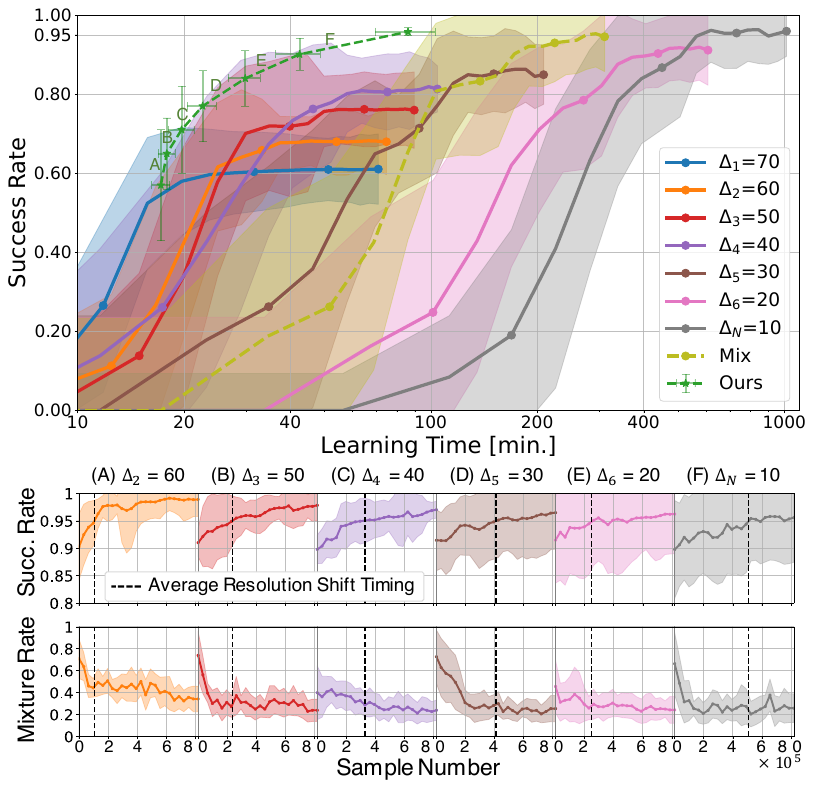}
    \vspace{-6mm}
    \caption{
        Comparison of learning time (top) and policy mixture rate (bottom):
        \textbf{(Top)} This compares learning time between fixed resolution learning and PRPD. $\Delta=70, \cdots, \Delta=10$ refers to resolutions as \tabref{table:calculation_time_rate_list}, while ``Mix'' refers to mixed resolutions (simultaneous policy learning), both up to $400\times128\times128$ samples (circle points are plotted every $100\times128\times128$ samples). 
        The success rate is evaluated 100 times in $\Delta=10$ at each iteration.
        \textbf{(Bottom)} This shows the task success rate and mixture rate transitions of PRPD in each resolution. The dashed line indicates when the scheduler changes resolution by achieving the target success rate $\hat{\tau}$. $\Delta=70$ is the initial environment and lacks a previous policy, so the mixture rate is omitted. Each curve plots the mean (and variance) over five experiments. 
    }
    \label{fig:learning_time}
\end{figure}

\begin{figure*}[t]
    \centering    \includegraphics[width=1.99\columnwidth]{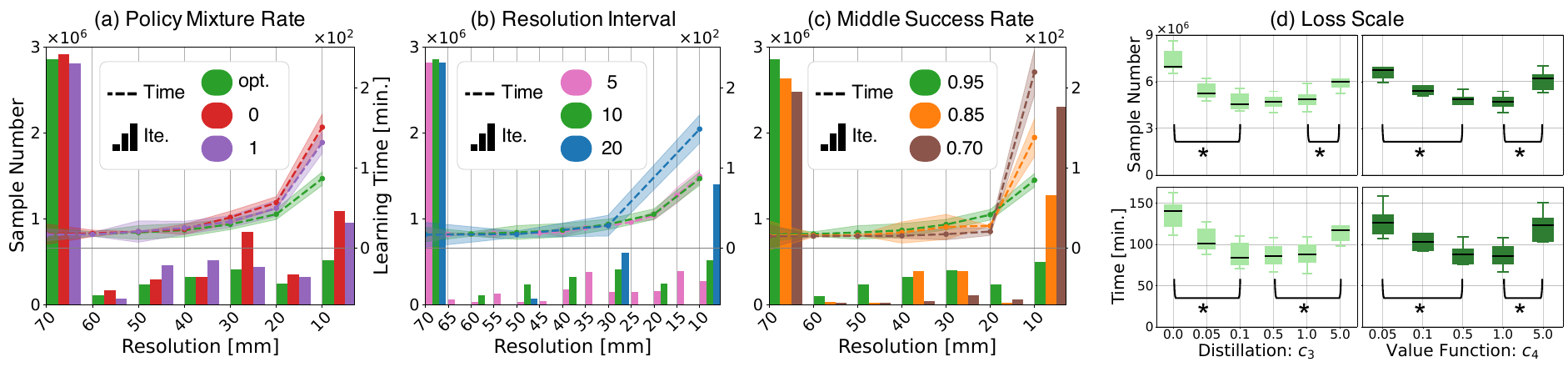}
    \caption{
        Summaries of analysis: 
        \textbf{(Left)} Performance comparison of PRPD components for (a) different patterns of policy mixture rate, (b) different numbers of grid interval, and (c) different numbers of target middle success rate.
        The box plots represent the sample numbers at which the policies achieve the target success rate $\hat{\tau}$ for each resolution.
        The dashed lines denote the total learning time from the initial resolution to the plotted resolution.
        $\alpha$ denotes the policy mixture rate, which is dynamically updated only in $\alpha=\text{opt.}$.
        \textbf{(Right)} (d) different scaling parameters of loss function in policy learning required to reach $\hat{\tau}$.
        Each curve and point of (a), (b), (c), and (d) plots the mean and variance per sample over five experiments (each box plot plots the mean).
        The learning sample is the value until the target success rate $\hat{\tau}$ is reached at the corresponding resolution (for Ours in (d), it is the summation value until $\hat{\tau}$ is reached at the final resolution $\Delta=10$). \textasteriskcentered \ mean $p < 0.05$.
    }
    \label{fig:alpha}
    \label{fig:interval}
    \label{fig:middle_threshold}
    \label{fig:divide-partition}
    \label{fig:time_sample}
    \label{fig:picture_sum}
\end{figure*}

\begin{figure}[t]
    \centering
    \includegraphics[width=0.95\columnwidth]{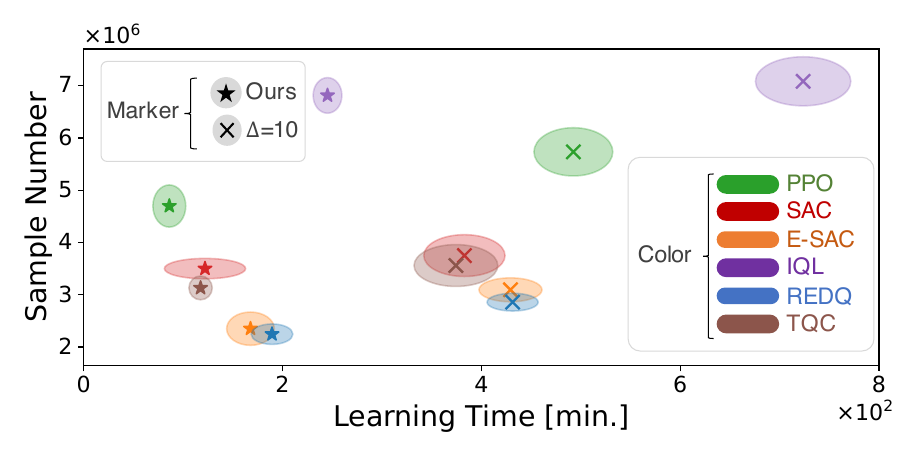}
    \caption{
        Relationship between learning time and number of samples in policy learning:
        Each point plots the mean and variance per sample over five experiments.
        Sample number is the summation value until $\hat{\tau}$ is reached at the final resolution $\Delta=10$.
    }
    \label{fig:compare_method}
\end{figure}

\subsection{Effect of Resolution Scheduling for Short-time Learning}
\label{ex-time}
    \subsubsection{Settings}
        PRPD schedules simulation resolution for short-time policy learning in fine-resolution simulations. 
        To verify its effectiveness, we compare the learning time and performance between PRPD and the policy learning conducted in a fine-resolution simulation.
        Additionally, policy learning in only coarse-resolution simulations may achieve high performance in a shorter time. Thus, we evaluate policy learning with fixed coarse-resolutions in various patterns of resolutions. In addition, policy learning in various resolutions simultaneously may be more efficient than scheduling resolutions. This framework is also compared.

    \subsubsection{Results}

        As shown in \figref{fig:learning_time}, PRPD learns control policies in less than one-seventh of the time needed to reach the highest success rate compared to fixed resolution learning. Coarser resolutions result in faster convergence, with a 20-fold difference in learning time between $\Delta=70$ and $\Delta=10$, but they also reduce performance by about \SI{35}{\%}. 
        Also, \tabref{table:compare_achieve_corr_delta10} shows that each policy achieves over \SI{95}{\%} success rate when evaluated at its own training resolution, while the success rate at $\Delta=10$ gradually decreases as the training resolution becomes coarser.
        PRPD converges in less than one-fourth of the time compared to simultaneous learning across all resolutions. These results demonstrate that the proposed resolution scheduling framework achieves the fastest policy learning.

\subsection{Effect of Conservative Policy Transfer}
\label{ex-MI}
    \subsubsection{Settings}
        PRPD applies a conservative policy transfer scheme to stabilize policy transfers caused by progressive resolution shifts.
        Specifically, the learning is stabilized by dynamically optimizing the mixture rate $\alpha$ (\equref{eq:proposed-alpha-update}).
        To evaluate the effectiveness of optimizing $\alpha$, we compared its performance with other baselines with constant $\alpha$.

    \subsubsection{Results}
        As shown in \figref{fig:alpha}, $\alpha=\text{opt.}$ achieves the shortest learning time compared to fixed mixture rates. Additionally, $\alpha=1$, where previous policies $\pi^{(n-1)}$ are used without optimization, outperforms $\alpha=0$, which does not use previous policies. These results confirm that the conservative policy transfer scheme enables learning with fewer samples.

        From the transition of the policy mixture rate $\alpha$ in PRPD shown in \figref{fig:learning_time}, $\alpha$ tends to be high in the early learning iterations of each resolution. This denotes that, during the initial phase of progressive resolution shifts, the update of the current policy $\pi^{(n)}$ is unstable and stronger regularization is added to prevent extensive updates from the previous policy $\pi^{(n-1)}$. As learning progresses, the update of $\pi^{(n)}$ stabilizes and the regularization decreases.

\begin{table*}[t]
    \begin{center}
        \caption{
            Sim-to-real experiment of rock excavation: 
            {\normalfont This experiment evaluated nine types of rocks (five differently shaped wooden blocks and three different-sized 3D-printed artificial rocks as shown in rock images) on three types of ground (for each element, three numbers correspond to each environment, silica sand, kinetic sand, and plastic pellets in order). ``Scale'' indicates approximate rock size. The first-row block shows ``Ours'' (PRPD) and ``w/o DR'' (PRPD without DR). The second-row block shows PPO at a fixed resolution. Each value indicates the success number per 20 trials, with the last column showing the average rate. The 20 trials test five policies at four positions (0, 60, 120, and 180 degrees) on a 100-mm radius circumference relative to the bucket's initial position. Task success is defined as rocks being held in the bucket and the bucket being off the ground.
            }
        }
        \label{table:achievement}
        \footnotesize
        \begin{tabular}{lcccccccccc}
            \toprule
                ID & No. 1 & No. 2 & No. 3 & No. 4 & No. 5 & No. 6 & No. 7 & No. 8 & No. 9 &  \\
                Rock Image &
                \begin{minipage}{0.13\columnwidth}
                  \scalebox{0.25}{\includegraphics{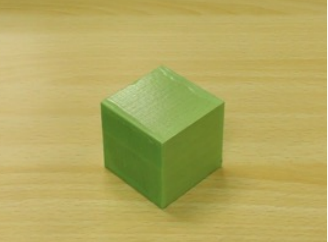}}
                \end{minipage} &
                \begin{minipage}{0.13\columnwidth}
                  \scalebox{0.25}{\includegraphics{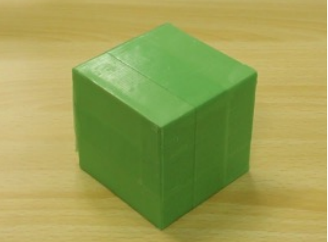}}
                \end{minipage} & 
                \begin{minipage}{0.13\columnwidth}
                  \scalebox{0.25}{\includegraphics{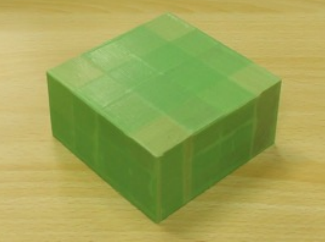}}
                \end{minipage} &
                \begin{minipage}{0.13\columnwidth}
                  \scalebox{0.25}{\includegraphics{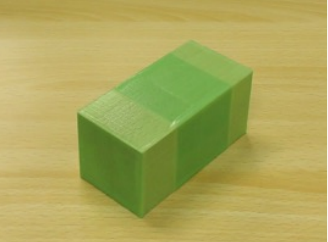}}
                \end{minipage} &
                \begin{minipage}{0.13\columnwidth}
                  \scalebox{0.25}{\includegraphics{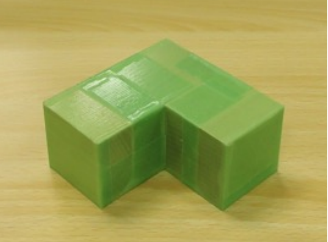}}
                \end{minipage} & 
                \begin{minipage}{0.13\columnwidth}
                  \scalebox{0.25}{\includegraphics{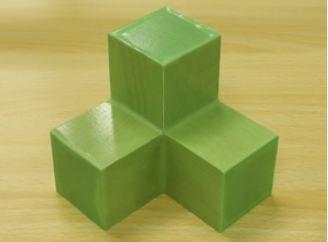}}
                \end{minipage} &
                \begin{minipage}{0.13\columnwidth}
                  \scalebox{0.25}{\includegraphics{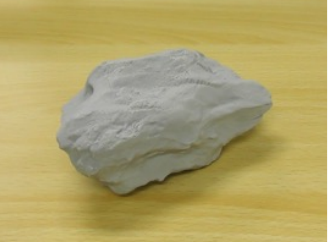}}
                \end{minipage} &
                \begin{minipage}{0.13\columnwidth}
                  \scalebox{0.25}{\includegraphics{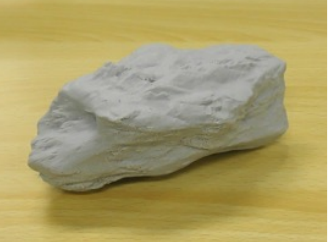}}
                \end{minipage} & 
                \begin{minipage}{0.13\columnwidth}
                  \scalebox{0.25}{\includegraphics{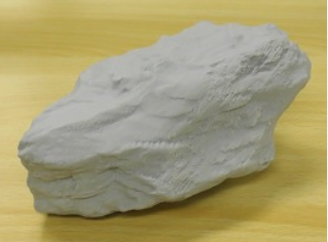}}
                \end{minipage} & 
                    Ave. \\
                Scale [mm$^3$] & $_{50\!\times\!50\!\times\!50}$ & $_{75\!\times\!75\!\times\!75}$ & $_{100\!\times\!100\!\times\!50}$ & $_{100\!\times\!50\!\times\!50}$ & $_{100\!\times\!100\!\times\!50}$ & $_{100\!\times\!100\!\times\!50}$ & $_{140\!\times\!70\!\times\!70}$ & $_{160\!\times\!80\!\times\!80}$ & $_{180\!\times\!90\!\times\!90}$ &  [\%] \\
            \midrule
                \textbf{Ours} & 19-19-20& 17-14-16& 18-18-15& 20-19-20& 19-17-19& 18-17-14& 19-20-18& 18-16-16& 18-15-17& 88.1 \\
                w/o DR & 16-14-18& 11-11-9& 15-13-14& 16-16-17& 14-13-12& 5-4-7& 12-10-6& 9-6-3& 3-4-1& 51.7 \\
                \midrule
                $\Delta=10$ & 20-19-19& 19-15-17& 18-17-18& 20-19-19& 18-16-17& 17-15-16& 18-19-18& 19-18-16& 18-16-17& 88.5 \\
                $\Delta=20$ & 19-16-20& 20-13-20& 17-14-16& 19-17-18& 18-14-20& 16-15-12& 19-16-18& 18-14-17& 15-14-14& 83.1 \\
                $\Delta=30$ & 19-17-18& 19-12-19& 16-16-15& 19-20-19& 17-15-16& 15-15-12& 17-16-14& 18-15-16& 13-12-10& 79.6 \\
                $\Delta=40$ & 19-19-19& 16-10-17& 17-18-16& 18-19-16& 18-15-16& 13-14-13& 18-15-15& 20-12-18& 11-11-9& 78.1 \\
                $\Delta=50$ & 20-18-20& 17-11-15& 14-11-12& 18-19-19& 14-11-15& 14-15-9& 17-14-14& 16-13-13& 12-11-11& 72.8 \\
                $\Delta=60$ & 17-15-17& 15-12-15& 13-11-10& 20-18-18& 16-13-17& 10-9-7& 19-17-17& 15-9-11& 6-5-6& 66.3 \\
                $\Delta=70$ & 18-18-20& 15-10-16& 14-9-13& 17-16-19& 16-12-15& 9-10-4& 18-16-16& 14-9-13& 3-3-4& 64.3 \\
            \bottomrule
        \end{tabular}
    \end{center}
\end{table*}

\subsection{Influence of Different Resolution Scheduling}
\label{ex-scheduling}

    \subsubsection{Settings}
        Since PRPD defines the resolution scheduler deciding learning resolution in a certain resolution interval $\Delta_{\mathcal{R}}$, the interval must be small enough to ensure a successful policy transfer.
        Also, as the resolution scheduler decides target success rates $\hat{\tau}$ for shifting resolution, the target middle success rates, the ones used except for the final resolution, may influence sample efficiency since middle tasks are not needed to solve perfectly. 
        Thus, we investigate the performance of PRPD with various $\Delta_{\mathcal{R}}$ and $\tau$. 
        
    \subsubsection{Results}
        As shown in \figref{fig:interval}, when the interval is coarse, the sample efficiency is poor, requiring about 1.5 times more samples. However, making the interval finer does not always improve efficiency, as there is almost no difference in sample efficiency when $\Delta_{\mathcal{R}}=10$ or less. This indicates that there is an optimal interval range where policy transfer works effectively.
        As shown in \figref{fig:middle_threshold}, lowering the target middle success rate $\tau$ requires more samples. This suggests that policy transfer is ineffective before the policy acquires sufficient skills.

\subsection{Impact of Loss Term Balancing in PRPD}
\label{ex-loss_scale}

    \subsubsection{Settings}
        To evaluate the robustness of the proposed PRPD framework with respect to loss weighting, we conduct a sensitivity analysis on the scale of each loss term in the overall objective as shown in Eq. (10). In this equation, $c_3$ and $c_4$ control the relative importance of the distillation loss and Q-value loss, respectively. This experiment systematically varies $c_3$ and $c_4$ to examine how loss balance affects training stability and performance.
        
    \subsubsection{Results}
        As shown in \figref{fig:picture_sum}, the proposed framework achieves both sample- and time-efficient learning across a broad range of loss weight settings. Sample efficiency is highest when $c_3$ lies between 0.1 and 1, suggesting that moderate incorporation of prior policy guidance is beneficial, whereas excessive emphasis may hinder learning. For $c_4$, optimal performance is observed when the value ranges from 0.5 to 1, indicating that $Q$-value estimation supports conservative transfer, though overly large weights may destabilize training. While the framework remains robust to moderate variations, setting $c_3$ or $c_4$ far outside these ranges can increase the required number of samples by approximately 1.5 times.

\subsection{Tradeoff between Time-Efficiency and Sample-Efficiency}
\label{ex-time_sample}

    \subsubsection{Settings}
        As mentioned in \chapref{related_work:sample_efficient_RL}, various methods have improved RL's sample efficiency, but less focus has been placed on ``time efficiency.''  This section shows that the proposed method outperforms traditional sample-efficient methods in time efficiency. 
        We evaluate Soft Actor-Critic (SAC) \cite{SAC} and its ensemble-based extension E-SAC \cite{RL_ensemble_auxiliary}, along with three recent off-policy algorithms recognized for their sample efficiency and strong empirical performance: Implicit Q-Learning (IQL) \cite{IQL}, Randomized Ensembled Double Q-learning (REDQ) \cite{REDQ}, and Truncated Quantile Critics (TQC) \cite{TQC}.
        These methods are tested within both the proposed PRPD framework and a fixed-resolution setting with $\Delta = 10$.

    \subsubsection{Results}
        As shown in \figref{fig:compare_method}, the proposed PRPD framework outperforms fixed resolution $\Delta = 10$ in time efficiency across all RL methods, despite being less sample-efficient. PPO, with the highest time efficiency among RL algorithms, achieved the fastest calculation time with PRPD. Under $\Delta = 10$, PPO had the longest calculation time due to its low sample efficiency. Other sample-efficient methods required less time and fewer samples than PPO, as the slow simulation time at $\Delta = 10$ increased sample collection time, making sample-efficient methods more time-efficient.

\subsection{Sim-to-Real Policy Transfer}
\label{ex-various_rock}

    This section validates the sim-to-real policy transfer across diverse real-world rock-excavation environments, including silica sand (low viscosity), kinetic sand (high viscosity), and plastic pellets (blocky shape and large grain size).
    \tabref{table:achievement} shows the evaluation of the transfer policies.
    The experiment shows that the success rate increases as the resolution improves in the real-world environment for fixed resolution policies, as was demonstrated with the simulation results. PRPD achieves performance comparable to fine-resolution fixed policy learning, indicating no adverse effects from using coarse-resolution simulations and successful sim-to-real transfer. 
    We also evaluated the differences in policy performance during sim-to-real transfer depending on the presence of DR parameters. \tabref{table:achievement} shows that policy performance dropped to approximately \SI{50}{\%} without DR. This indicates a reality gap between the simulation and real-world environments, mitigated by applying DR.
    These results demonstrate that the variable-resolution rock excavation simulator and the proposed policy learning framework achieve approximately \SI{90}{\%} success rates for nine types of real rock environments.

\section{Discussions}
\label{s:dis}

        \textbf{Finer resolution assumption:} 
            In this paper, we assumed that finer resolution in excavation simulators leads to better transfer policy performance in real-world environments, supported by the experimental results in \chapref{ex-various_rock}. However, this assumption may not always hold. The relationship between resolution and the reality gap could reverse or remain unchanged beyond a certain point. While finer resolutions may reveal physical differences, these might not significantly impact policy learning. The influence of policy generalization through DR on this relationship is also considered. These discussions are left for future work.

        \textbf{Automatic determination of resolution interval:} 
            As an extension of the proposed PRPD, automatic determination of resolution intervals $\Delta_{\mathcal{R}}$ could be considered. In \chapref{ex-scheduling}, we showed that finer $\Delta_{\mathcal{R}}$ allows for learning with fewer samples. Since these parameters depend on the task and policy learning algorithm, determining a single optimal value is difficult. However, developing a framework to automatically adjust simulation intervals based on their relationship would be beneficial.

        \textbf{Applicability of proposed framework:} 
            This paper focuses on rock excavation within earthwork tasks in the development of a simulator and an effective learning algorithm. The proposed RL framework of progressively changing simulation resolution for faster policy learning could be extended to other excavation tasks or the installation of non-rock objects. Additionally, this approach could broadly apply to tasks where simulation resolution affects simulation time beyond just earthwork tasks.
            This paper demonstrated a 7-fold improvement in learning time efficiency (Ours: 90 minutes, fixed highest resolution: 600 minutes), resulting in an 8-hour time gap. This difference is likely to become more significant with increasing state-action space complexity or task difficulty. Furthermore, while the finest resolution in this study was set to $\Delta = 10$, targeting even higher resolutions would likely widen the gap in learning time.

        \textbf{Simulation Enginnering Cost:}
            In this paper, multi-resolution simulations are constructed as independent environments with different particle sizes, corresponding to varying spatial resolutions. Each resolution requires a separate simulation run with appropriately adjusted parameters to generate simulations at the corresponding resolution. We assume that such environments can be prepared in advance, and our framework is designed to efficiently learn high-resolution policies under this assumption. In practice, the engineering cost of creating variable-resolution simulators is relatively low in modern earthwork simulation platforms. For example, in particle-based simulators such as Vortex Studio \cite{vortex-DRL-sim2sim} and OPERA \cite{opera}, modifying a single configuration parameter is sufficient to control particle size and generate simulations at different resolutions.

\section{Conclusion}
\label{s:con}

    In this paper, we propose time-efficient RL framework PRPD to address the time inefficiency of a fine-resolution rock-excavation simulator. We evaluated PRPD by developing a variable-resolution rock-excavation simulator using Isaac Gym. PRPD significantly reduced policy learning time in the simulator. Additionally, the learned policy was successfully transferred to the real-world environment, robustly removing previously unseen rocks.

\addtolength{\textheight}{-0cm}   


\bibliographystyle{IEEEtran}
\bibliography{IEEEabrv,paper}

\end{document}